\documentclass[preprint,12pt]{elsarticle}

\usepackage{amssymb}

\usepackage{lineno}
\usepackage{hyperref}
\usepackage{subfig}
\usepackage[dvipsnames]{xcolor}
\usepackage[normalem]{ulem}

\journal{Neurocomputing}

\begin{document}

\begin{frontmatter}



\title{Benefiting from Multitask Learning to \\ Improve Single Image Super-Resolution}

\author[label1]{Mohammad Saeed Rad}
\ead{saeed.rad@epfl.ch}
\author[label1]{Behzad Bozorgtabar}
\author[label2]{Claudiu Musat}
\author[label2]{Urs-Viktor Marti}
\author[label2]{Max Basler}
\author[label3]{Haz{\i}m Kemal Ekenel}
\author[label1]{Jean-Philippe Thiran}

\address[label1]{Signal Processing Laboratory 5, \'Ecole Polytechnique F\'ed\'erale de Lausanne (EPFL), Lausanne, Switzerland}

\address[label2]{AI Lab, Swisscom AG, Lausanne, Switzerland}

\address[label3]{Istanbul Technical University, Istanbul, Turkey} 
\begin{abstract}
Despite significant progress toward super resolving more realistic images by deeper convolutional neural networks (CNNs), reconstructing fine and natural textures still remains a challenging problem. Recent works on single image super resolution (SISR) are mostly based on optimizing pixel and content wise similarity between recovered and high-resolution (HR) images and do not benefit from recognizability of semantic classes. In this paper, we introduce a novel approach using categorical information to tackle the SISR problem; we present a decoder architecture able to extract and use semantic information to super-resolve a given image by using multitask learning, simultaneously for image super-resolution and semantic segmentation. To explore categorical information during training, the proposed decoder only employs one shared deep network for two task-specific output layers. At run-time only layers resulting HR image are used and no segmentation label is required. Extensive perceptual experiments and a user study on images randomly selected from COCO-Stuff dataset demonstrate the effectiveness of our proposed method and it outperforms the state-of-the-art methods.
\end{abstract}


\end{frontmatter}
\section{Introduction}
\label{sec:intro}
Single image super-resolution (SISR) has many practical computer vision applications \cite{paper_application_0, paper_application_1, paper_application_2, paper_application_3}, which aims at recovering high-resolution (HR) images from a set of prior examples of paired low-resolution (LR) images. Although many SISR methods have been proposed in the past decade, recovering high-frequency details and realistic textures in a plausible manner are still challenging. Having said that, this problem is ill-posed, meaning each LR image might correspond to many HR images and the space of plausible HR images scales up quadratically with the image magnification factor.\\

\begin{figure}
\centering
\includegraphics[height=5cm]{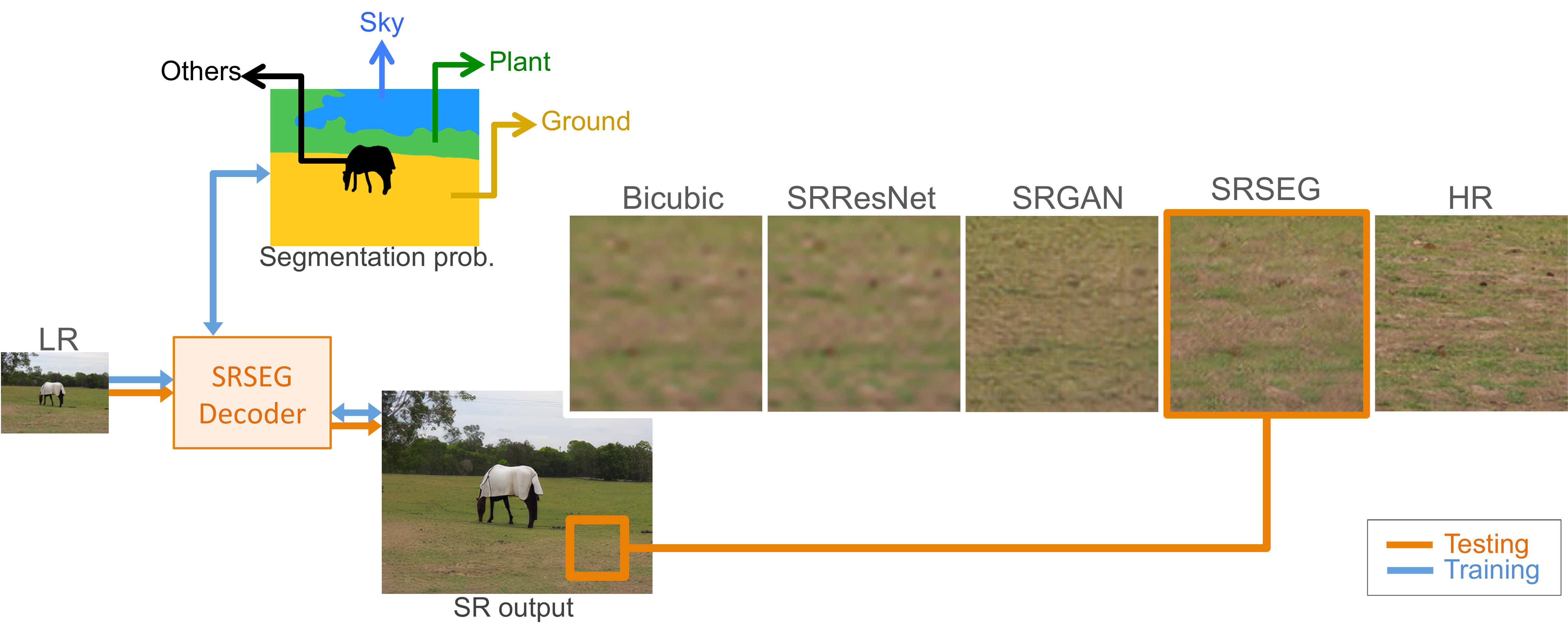}
\caption{The proposed single image super-resolution using multitask learning. This network architecture enables reconstructing SR images in a content-aware manner; during training (blue arrows), an additional objective function for semantic segmentation is used to force the SR to learn categorical information. At run-time we only reconstruct the SR image (orange arrows). In this work, we prove that learning semantic segmentation task in parallel with SR task can improve the reconstruction quality of SR decoder. Results from left to right: bicubic interpolation, SRResNet, SRGAN \cite{paper_tweeter_0}, and SRSEG (this work). Best viewed in color.}
\label{fig:cover_image}
\end{figure}

To tackle such an ill-posed problem numerous deep learning methods have been proposed to learn mappings between LR and HR image pairs \cite{paper_deep_sr_0, kim2016, paper_deep_sr_2, paper_deep_sr_3}. These approaches use various objective functions in a supervised manner to reach the current state-of-the-art. Conventional pixel-wise Mean Squared Error (MSE) is the commonly used loss to minimize pixel-wise similarity of the recovered HR image and the ground truth in an image space. However, \cite{paper_mse_0, paper_tweeter_0} show that lower MSE does not necessarily reflect a perceptually better SR result. Therefore, \cite{paper_perc_0} proposed perceptual loss to optimize a SR model in a feature space instead of pixel space. Significant progress has been recently achieved in SISR by applying Generative Adversarial Networks (GANs) \cite{paper_tweeter_0, paper_srgan_2, paper_segmentation}. GANs are known for the ability to generate more appealing and realistic images and have been used in different image synthesis-based applications \cite{isola2017image, bozorgtabar2019using, mahapatra2018efficient, bozorgtabar2019learn}. 

\subsection{Does semantic information help?}
Despite significant progress toward learning deep models to super resolve realistic images, the proposed approaches still cannot fully reconstruct realistic textures; intuitively, it is expected to have a better reconstruction quality for common and known types of textures, e.g., ground soil and sea waves, but experiments show that the reconstruction quality is almost the same for a known and an unknown type of texture, e.g., a fabric with a random pattern. Although loss functions used in image super-resolution, e.g., perceptual and adversarial losses, generate appealing super-resolved images, they try to match the global level statistics of images without retaining the semantic details of the content. \cite{paper_segmentation} shows that variety of different HR image patches could have very similar LR counterparts, and as a consequence, similar SR images are reconstructed for categorically different textures using current state-of-the-art methods. They also prove that more realistic textures could be recovered by using an additional network to obtain prior knowledge and afterward use it as a secondary input in SR decoder. \\

In this work, we prove that a single SR decoder is capable of learning this categorical knowledge by using multitask learning. As \cite{paper_multitasklearning} emphasizes, multitask learning improves generalization by using the domain information contained in the training signals of related tasks. This improvement is the result of learning tasks in parallel while using a shared representation; in our case, what is learned for semantic segmentation task can help improving the quality of SR task and vice versa.\\

\subsection{Our contribution}

In this paper, we propose a novel architecture to reconstruct SR images in a content-aware manner, without requiring an additional network to predict the categorical knowledge. We show that this can be done by benefiting from multitask learning simultaneously for SR and semantic segmentation tasks. An overview of our proposed method is shown in Figure \ref{fig:cover_image}. We add an additional segmentation output in a way that the same SR decoder learns to segment the input image and generate a recovered image. We also introduce a novel boundary mask to filter out unrelated segmentation losses related to imprecise segmentation labels. The semantic segmentation task forces the network to learn the categorical knowledge. These categorical priors learned by the network are characterizing the semantic classes of different regions in an image and are the key to recover more realistic textures. Our approach outperforms quality of recovering textures of state-of-the-art algorithms in both qualitative and user studies manner. \\

\noindent
Our contributions can be summarized as follows:
\begin{itemize}
  \item We propose a framework that uses segmentation labels during training to learn a CNN-based SR model in a content-aware manner.
  
  \item We introduce a novel boundary mask to have an additional spatial control over categorical information within training examples and their segmentation label, and filter out their irrelevant information for SR task.
  
  \item Unlike existing approaches for content-aware SR, the proposed method does not require any semantic information at the test time. Therefore, neither segmentation label nor additional computation is required at test time while benefiting from categorical information.
  
  \item Our method is trained end-to-end and is easily reproducible.
  
  \item Our experimental results, including an extensive user study, prove the effectiveness of using multitask learning for SISR and semantic segmentation and show that SISR of high perceptual quality can be achieved by using our proposed objective function.
\end{itemize}

In the remainder of this paper, first, in Section~\ref{sec:relatedwork}, we review the related literature. Then, in Section~\ref{sec:method}, we give a detailed explanation about our design including the used dataset and our training parameters. In Section~\ref{sec:experiments} we present experimental results and computational time, and discuss the effectiveness of our proposed approach. Finally, we conclude the paper in Section~\ref{sec:conclusion} and also mention the future research directions.\\

\newpage
\section{Related Work}
\label{sec:relatedwork}
\subsection{Single image super-resolution}
SISR has been widely studied for decades and many different approaches have been proposed; from simple methods such as bicubic interpolation and Lanczos resampling \cite{paper_Lanczos}, to dictionary learning \cite{paper_Yang_dic} and self-similarity \cite{paper_self_similarity_1, Huang-CVPR-2015} approaches. With the advances of deep CNNs, the state-of-the-art SISR methods have been built based on end-to-end deep neural networks and achieved significantly superior performances, thus we only review relevant recent CNN-based approaches. 

An end-to-end CNN-based approach was proposed by \cite{dong2016} to learn the mapping of LR to HR images. The concept of residual blocks and skip-connections \cite{HeZR016,paper_deep_sr_2} were used by \cite{paper_tweeter_0} to facilitate the training of CNN-based decoders. A laplacian pyramid network was presented in \cite{lai2017} to progressively reconstruct the sub-band residuals of high-resolution images. The choice of the objective function plays a crucial role in the performance of optimization-based methods. These works used various loss functions; the commonly used loss term is the pixel-wise distance between the super-resolved and the ground-truth HR images for training the networks \cite{paper_mse_0,dong2016,kim2016,paper_rcan}. However, using those functions as the only optimization target leads to blurry super-resolved images due to the pixel-wise average of possible solutions in the pixel space. 

A remarkable improvement in terms of the visual quality in SISR is the so-called perceptual loss \cite{paper_perc_0}. This loss function benefits from the idea of perceptual similarity \cite{paper_mse_6} and seeks to minimize the distance loss over feature maps extracted from a pre-trained network, e.g., VGG \cite{paper_vgg}. In a similar work, \cite{mechrez2018learning} proposes contextual loss to generate images with natural image statistics, which focuses on the feature distribution rather than merely comparing the appearance.

More recently, the concept generative adversarial network (GAN) \cite{paper_gan} is used for image super-resolution task, which achieves state-of-the-art results on various benchmarks in terms of reconstructing more appealing and realistic images \cite{paper_mse_0, paper_tweeter_0, paper_segmentation}. The intuition behind its excellent performance is that GAN drives the image reconstruction towards the natural image manifold producing perceptually more convincing solutions.  Having said that, it also uses a discriminator to distinguish between the generated and the original HR images, which is found to produce more photo-realistic results.\\

\subsection{Super-resolution faithful to semantic classes}
Semantic information has been used in different studies for variant tasks; \cite{deblurring_Wenqi} proposed a method to benefit from semantic segmentation for video deblurring. For image generation, \cite{ChenK17aa} used semantic label to produce an image with photographic appearance. \cite{IsolaZZE16} used the same idea to perform image to image translation. The SISR method proposed by \cite{paper_segmentation} is more relevant to our work. They use an additional segmentation network to estimate probability maps as prior knowledge and use them in existing super-resolution networks. Their segmentation network is pre-trained on the COCO dataset \cite{lin2014microsoft} and then fine-tuned on the ADE dataset \cite{zhou2017scene}. They show that it is possible to recover textures faithful to categorical priors estimated through the pre-trained segmentation network, which generates intermediate conditions from the prior and broadcasts the conditions to the super-resolution network.

However, in this paper, we do not have an additional segmentation network, instead our super-resolution method is built on multitask end-to-end deep networks with the shared feature extraction parameters to learn semantic information. The intuition behind this proposed method is that the model can exploit features for both tasks, such a model, during training, is forced to explore categorical information while super-resolving the image. Therefore, the segmentation labels would be used only during the training phase and no additional segmentation labels would be required as the input at run-time.

\section{Multitask Learning for Image Super-Resolution}
\label{sec:method}

Our ultimate goal is to train a SISR in a multitask manner, simultaneously for image super-resolution and semantic segmentation. Our proposed SR decoder only employs one shared deep network and keeps two task-specific output layers during training to force the network learn semantic information. If the network converges for both tasks, we can be sure that the parameters of the shared feature extractor have explored categorical information while super-resolving the image. In this section we present our proposed architecture and the objective function used for training. We also introduce a novel boundary mask used to simplify the segmentation task.

\begin{figure}
\centering
\includegraphics[height=5.6cm]{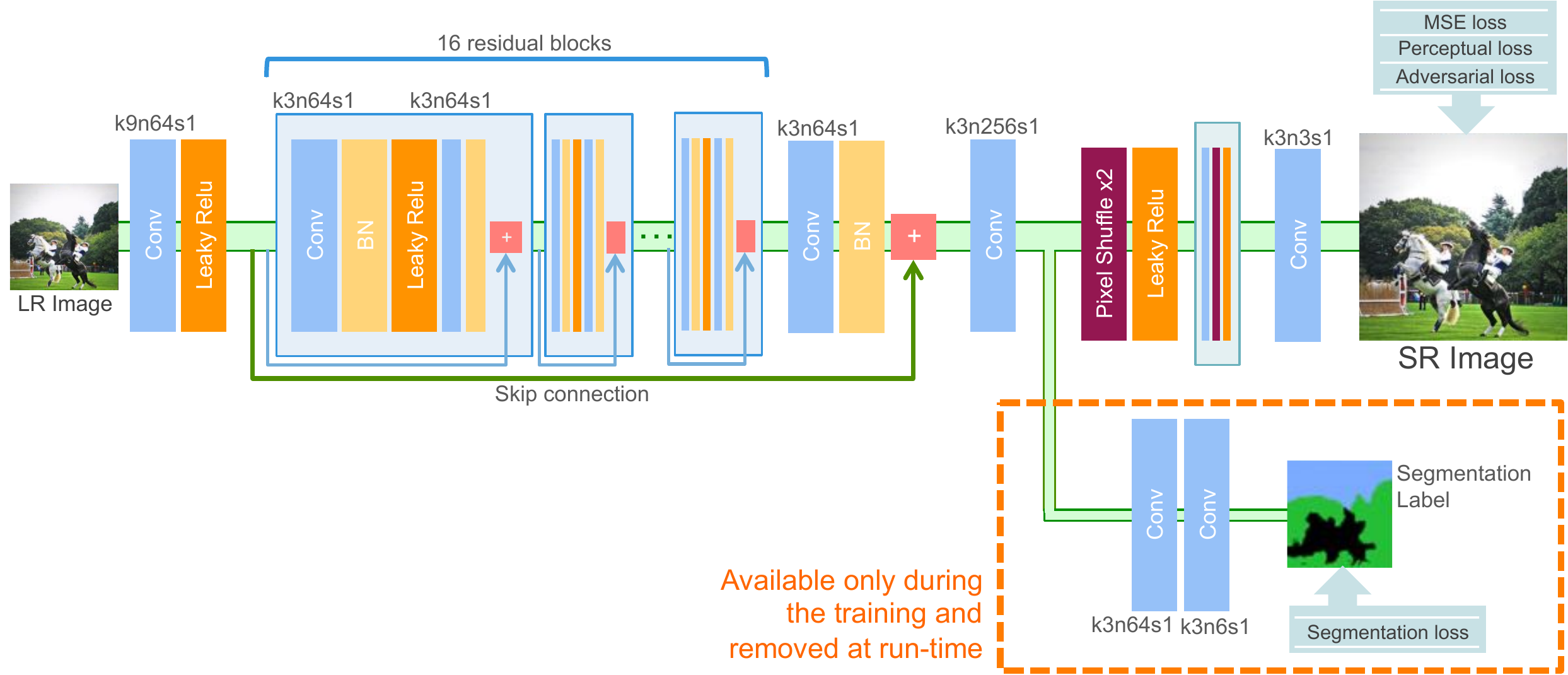}
  \caption{Architecture of the decoder. We train the SR decoder (upper part) in a multitask manner by introducing a segmentation extension (lower part). Feature extractor is shared between both super-resolution and segmentation tasks. The segmentation extension is only available during the training process and no segmentation label is used at the run-time. In this schema, $k$, $n$ and $s$ correspond respectively to kernel size, number of feature maps, and strides.}
\label{fig:system}
\end{figure}

\subsection{Architecture}
Figure \ref{fig:system} shows the multitask architecture used during training; the upper part (first row) shows SR generator, from the LR to HR image, while the lower part (second row) is the extension used to predict segmentation class probabilities. The role of segmentation extension layers of our design is to force the feature extractor parameters learn categorical information. These non-shared layers, generating segmentation probabilities, are not used during SR run-time. Each part is presented in more details as follows:

\begin{itemize}
  \item \textbf{SR generator} The generator network is a feed-forward CNN; the input image $I^{LR}$ is passed through a convolution block followed by LeakyReLU activation layer. The output is subsequently passed through 16 residual blocks with skip connections. Each block has two convolutional layers with $3\times3$ filters and $64$ feature maps, each one followed by a batch normalization and LeakyReLU activation. The output of the final residual block, concatenated with the features of the first convolutional layer, is inputted through two upsampling stages. Each stage doubles the input image size. Finally, the result is passed through a convolution stage to get the super-resolved image $I^{SR}$. In this study, we only investigate a scale factor of 4, but depending on the desired scaling, the number of upsampling stages can be changed. 
  \item \textbf{Segmentation extension} The segmentation extension uses the output of the SR generator feature extractor part, just before the first upsampling stage, and convert it to a segmentation probability by passing it through two convolutional layers. The computational complexity of this stage needs to be as limited as possible, as we wish that shared-layers with SR generator learn categorical information and not only layers from segmentation extension.
\end{itemize}

The parameters of the generator, for both segmentation and SR tasks, are obtained by minimizing the $\mathcal{L}_{total}$ loss function presented in Section \ref{subsec:loss}. This loss function consists also of a GAN \cite{paper_gan}-based adversarial loss, which requires a discriminator network. This network discriminates real HR images from generated SR samples. We define our discriminator architecture similar to \cite{paper_tweeter_0}; it consists of multiple convolutional layers with the kernels increasing by a factor of $2$ from $64$ to $512$. We use Leaky ReLU and strided convolutions to reduce the image dimension while doubling the number of features. The resulting $512$ feature maps are followed by two dense layers. Finally, the image is classified as real or fake by a final sigmoid activation function.

\begin{figure}
\centering
\subfloat[]{{\includegraphics[height=3cm]{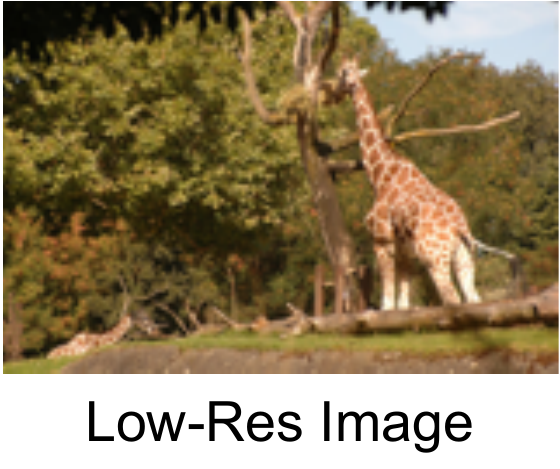} }}
\qquad \qquad \subfloat[]{{\includegraphics[height=3cm]{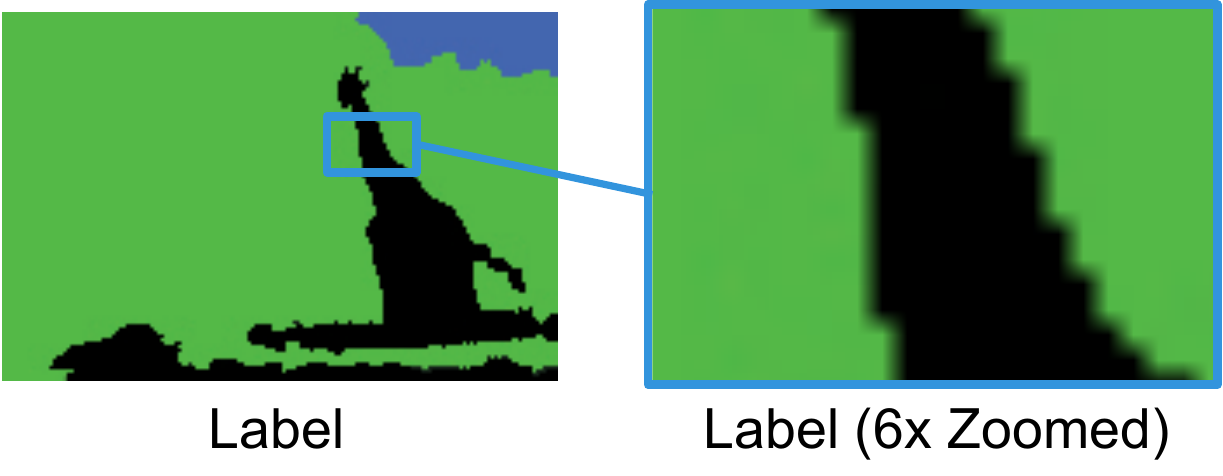} }}
\caption{An example showing the accuracy and resolution of a pixel-wise semantic segmentation label (b) of a low resolution image (a). As both segmentation and super-resolution networks share layers, the inaccurate segmentation labels result inaccurate edges in super-resolved images.}
\label{fig:segmentation_accuracy}
\end{figure} 

\subsection{Boundary mask}
\label{subsec:boundary_mask}
Although segmentation labels of available datasets, e.g., \cite{paper_coco_stuff}, to be used for segmentation task, are created by an expensive labeling effort, they still lack of precision close to boundaries of different classes as can be seen in Figure~\ref{fig:segmentation_accuracy}. Our experiments show that as shared features are used for generating the SR image and segmentation probabilities, this lack of boundaries' precision in segmentation labels affects the edges in the SR image too. Therefore, we use a novel boundary mask ($M_{boundary}$) to filter out any segmentation losses from areas close to object boundaries from training images.

\begin{figure}[h]
\centering
\includegraphics[height=2.4cm]{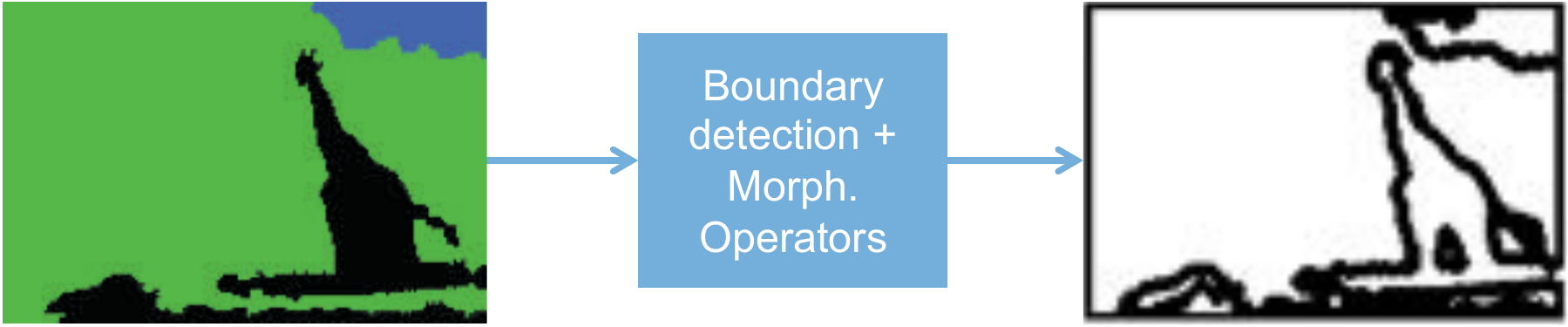}
  \caption{The boundary mask generation. The black pixels of the results represent areas close to the edges while white pixels could be either background or foreground.}
\label{fig:segmentation_mask}
\end{figure}

In order to generate such a boundary mask, first, we calculate the derivative of the segmentation label to get the boundaries of different classes in the low resolution image. Then, we compute the dilation of results with a disk of size $d_1$ to create a thicker strip around edges of each class. An example of converting the segmentation label to the boundary mask is shown in Figure~\ref{fig:segmentation_mask}. In Section~\ref{sec:experiments} the effectiveness of using such boundary masks is shown.

\vspace{-1mm}
\newcommand{\calL}{\mathcal{L}}
\subsection{Loss function}
\label{subsec:loss}
We define the $\mathcal{L}_{total}$ as a combination of pixel-wise loss ($\calL_{MSE}$), perceptual loss ($\calL_{vgg}$), adversarial loss ($\calL_{adv}$), and segmentation loss ($\calL_{seg}$) filtered by our novel boundary mask ($M_{boundary}$) presented in Section~\ref{subsec:boundary_mask}. The overall loss function is given by: 

\begin{equation}
\calL_{total} = \alpha \calL_{MSE} + \beta \calL_{vgg} + \gamma \calL_{adv} + \delta M_{boundary}. \calL_{seg}
\label{eq:cf2}
\end{equation}

where $\alpha$, $\beta$, $\gamma$, and $\delta$ are the corresponding weights of each loss term used to train our network. In the following, we present each term in detail:

\begin{itemize}
  \item \textbf{Pixel-wise loss} The most common loss in SR is the pixel-wise Mean Squared Error (MSE) between the original image and the super-resolved image in the image space \cite{paper_mse_0,paper_tweeter_0,paper_deep_sr_0}. However, using it alone mostly results in finding pixel-wise averages of plausible solutions, which seems over-smoothed with poor perceptual qualities and lack of high-frequency details such as textures \cite{paper_mse_4,paper_mse_5,paper_mse_6}.
   \item \textbf{Perceptual loss} \cite{paper_mse_0} and \cite{paper_tweeter_0} used the idea of measuring the perceptual similarity by computing the distance of feature spaces of the images. First, both HR and SR images are mapped into a feature space by a pre-trained model, VGG-16 \cite{paper_vgg} in our case. Then, the perceptual loss is calculated by the $L2$ distance and using all $512$ feature maps of ReLU 4-1 layer of the VGG-16.
   \item \textbf{Adversarial loss} Inspired by \cite{paper_tweeter_0} we add the discriminator component of the mentioned GAN architecture to our design. This encourages our SR decoder to favor solutions that resolve more realistic and natural images, by trying to trick the discriminator network. It also results perceptually superior solutions to solutions obtained by minimizing pixel-wise MSE and perceptual loss.
   \item \textbf{Segmentation loss} While using segmentation for SR application is new for the community, semantic segmentation as a stand-alone task has been investigated for years. The most commonly used loss function for the task of image segmentation is a pixel-wise cross entropy loss (or log loss) \cite{cross_entropy_1, cross_entropy_2, cross_entropy_3}. In this work, we also use the cross entropy loss function to examine each pixel individually and compare the class predictions (depth-wise pixel vector) to the one-hot encoded label; it measures the performance of a pixel-wise classification model whose output is a probability value between zero and one for each pixel and category.
\end{itemize}

\subsection{Dataset}
Training the proposed network in a supervised manner requires a considerable number of training examples with ground-truths for both semantic segmentation and super resolution tasks. Therefore the choices of datasets are limited to the ones with available segmentation labels. We use a random sample of 60 thousand images from the COCO-Stuff database \cite{paper_coco_stuff}, which contains semantic labels for 91 stuff classes for segmentation task. We only choose images from five main background classes to be able to focus on texture quality and prove the concept: sky, ground, buildings, plants, and water. Each one of them contains multiple sub classes in COCO-Stuff dataset, e.g., water contains seas, lakes, rivers, etc. and plants contain trees, bushes, leaves, etc., but in this work we consider them as a single class. Any other object or background existing in an image is labeled as "others" (the sixth class). More than 12 thousand images from each category were used to train our network. We obtained the LR images for the SR task by downsampling the HR images of the same database using the MATLAB imresize function with the bicubic kernel and downsampling factor 4 (all experiments were performed with a scaling factor of ×4). For each image, we crop a random $82 \times 82$ HR sub image for training.\\

\subsection{Training and parameters}
In order to successfully converge to parameters compatible for both SR and the segmentation task, the training was done in different steps; first, the generator was trained for 25 epochs with only pixel-wise mean squared error as the loss function. Then the segmentation loss function was added and training continued for 25 more epochs. Finally, the loss function presented in Section \ref{subsec:loss} (including adversarial and perceptual losses) was used for 55 more epochs. The weights of each term in loss function presented in Eq. \ref{eq:cf2} were chosen as follows: as proposed by \cite{paper_tweeter_0}, $\alpha$, $\beta$, and $\gamma$ were respectively fixed to $1.0$, $2 \times 10^{-6}$, and $1 \times 10^{-3}$. $\delta$ were tuned and fixed to $0.8$. The Adam optimizer \cite{paper_adam} was used for all the steps. The learning rate was set to $1 \times 10^{-3}$ and then was decayed by a factor of 10 every 20 epochs. We also alternately optimized the discriminator with the setting proposed by \cite{paper_tweeter_0}.

\begin{figure}
\centering
\includegraphics[height=2.3cm]{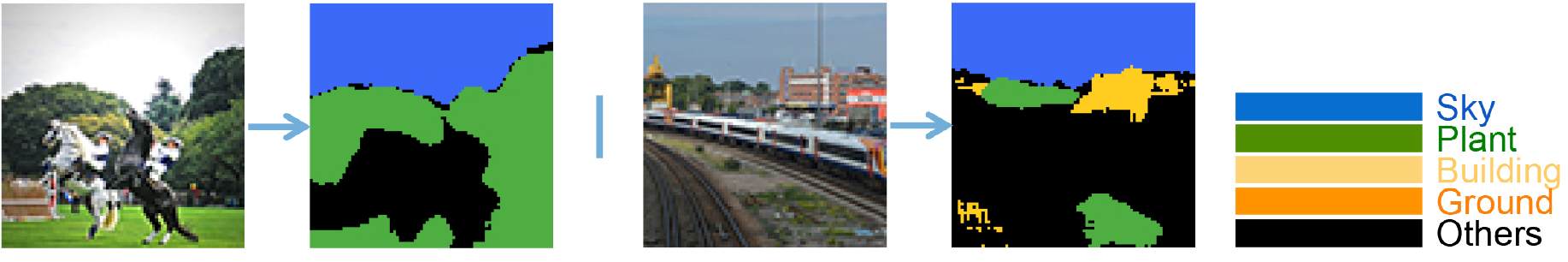}
\caption{Two examples of segmentation prediction results. The artifacts close to boundaries (imprecise edges and black strips around them) are the result of applying boundary mask in a way that the generator does not focus on class probabilities around boundaries and have a random prediction on those areas.}
\label{fig:segmentation_res}
\end{figure}

As explained previously, to not consider a segmentation prediction error close to boundaries of objects/backgrounds, the segmentation loss is filtered by a boundary mask as introduced in Section~\ref{subsec:boundary_mask}. Figure \ref{fig:segmentation_res} shows the segmentation prediction results of two training images; the artifacts close to boundaries (imprecise edges and black strips around them) are the result of applying a boundary mask. This mask makes the network not consider the class probabilities around boundaries and have a random prediction on those areas.\\

\section{Results and Discussion}
\label{sec:experiments}
In this section, we first investigate the effectiveness of using the presented boundary mask in the proposed approach. Then, we evaluate and discuss the benefits of introducing multitask learning for SR task by performing qualitative experiments, an extensive user study, and an ablation study. Finally, we discuss the computational time of the proposed approach.


\subsection{Effectiveness of boundary masks}
\begin{figure}
\centering
\subfloat[]{{\includegraphics[height=2.8cm]{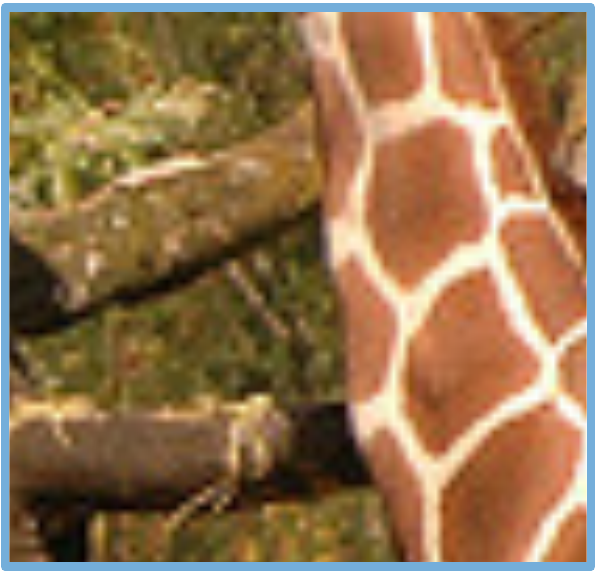} }}
\subfloat[]{{\includegraphics[height=2.8cm]{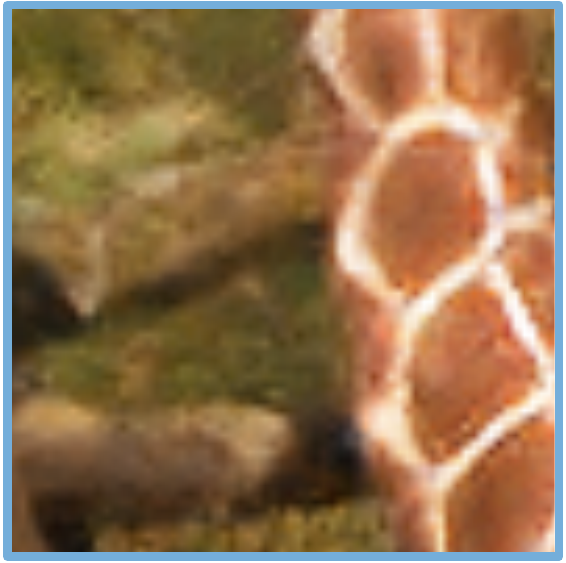} }}
\subfloat[]{{\includegraphics[height=2.8cm]{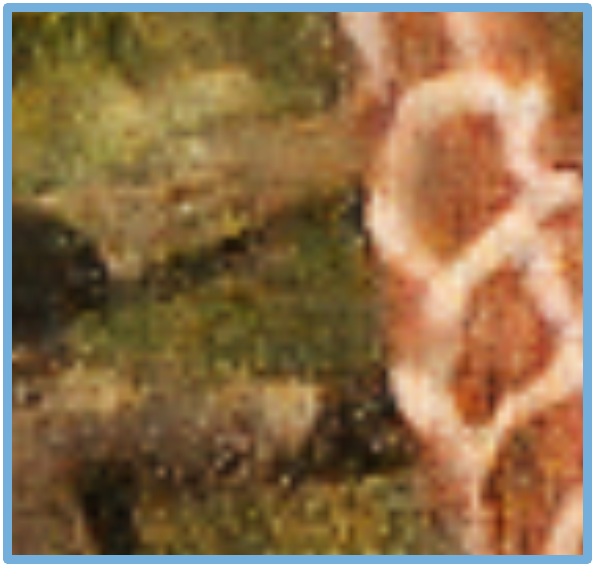} }}
\subfloat[]{{\includegraphics[height=2.8cm]{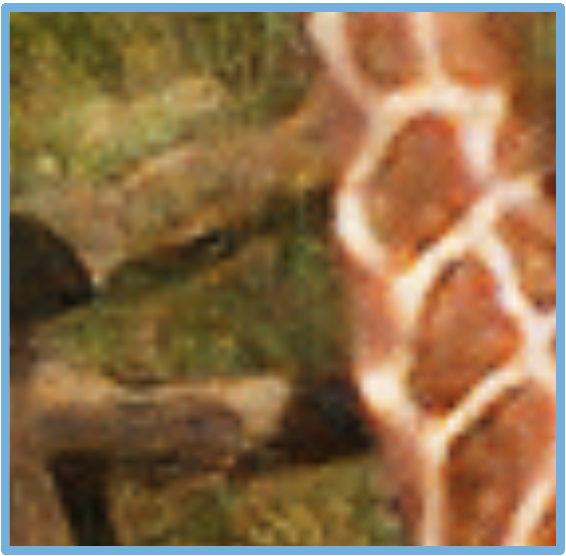} }}
\caption{(a) Ground-truth, (b) SRGAN, (c) SRSEG, (d) Masked-SRSEG. While SRGAN still has the most accurate edges in this example, both masked and unmask SRSEG network constructs more realistic textures in the background and are closer to ground-truth. All images are cropped from Figure \ref{fig:segmentation_accuracy}.a and zoomed by a factor of 6 ($6 \times$).}
\label{fig:seg_0}
\end{figure}

As explained previously in Section \ref{subsec:boundary_mask} in this work we use a novel boundary mask ($M_{boundary}$) to filter out all segmentation losses from areas close to object boundaries during training. The goal of this masking is to avoid forcing SR network to learn imprecise boundaries existing in segmentation labels. Figure \ref{fig:seg_0} shows the SR results comparing the effect of segmentation mask; comparing Figure \ref{fig:seg_0}.c to \ref{fig:seg_0}.d shows the improvement in reconstructing sharper edges using segmentation with mask rather than without mask. In this example, both Figures \ref{fig:seg_0}.c and \ref{fig:seg_0}.d have the closest textures to the ground-truth comparing to Figure \ref{fig:seg_0}.b, however, the object in the super-resolved image without using segmentation information has the sharpest edges; this can be explained by the fact that we only considered background categories (``sky'', ``plant'', ``buildings'', ``ground'', and ``water'') because of their specific appearance and to prove the concept. All type of objects, e.g., giraffe in this example, are included in ``Other'' category, therefore, no specific pattern is expected to be learnt for this category. As a future work, more object categories can be added to the training examples.

\subsection{Qualitative results}
Standard benchmarks such as Set5 \cite{paper_set5}, Set14 \cite{paper_set14}, and BSD100 \cite{paper_bsd100} mostly do not contain the background categories studied in this research, therefore, first we evaluate our method on a test set consisting of random images of the COCO-stuff dataset \cite{paper_coco_stuff}.

\begin{figure}
\centering
\includegraphics[width=14.0cm]{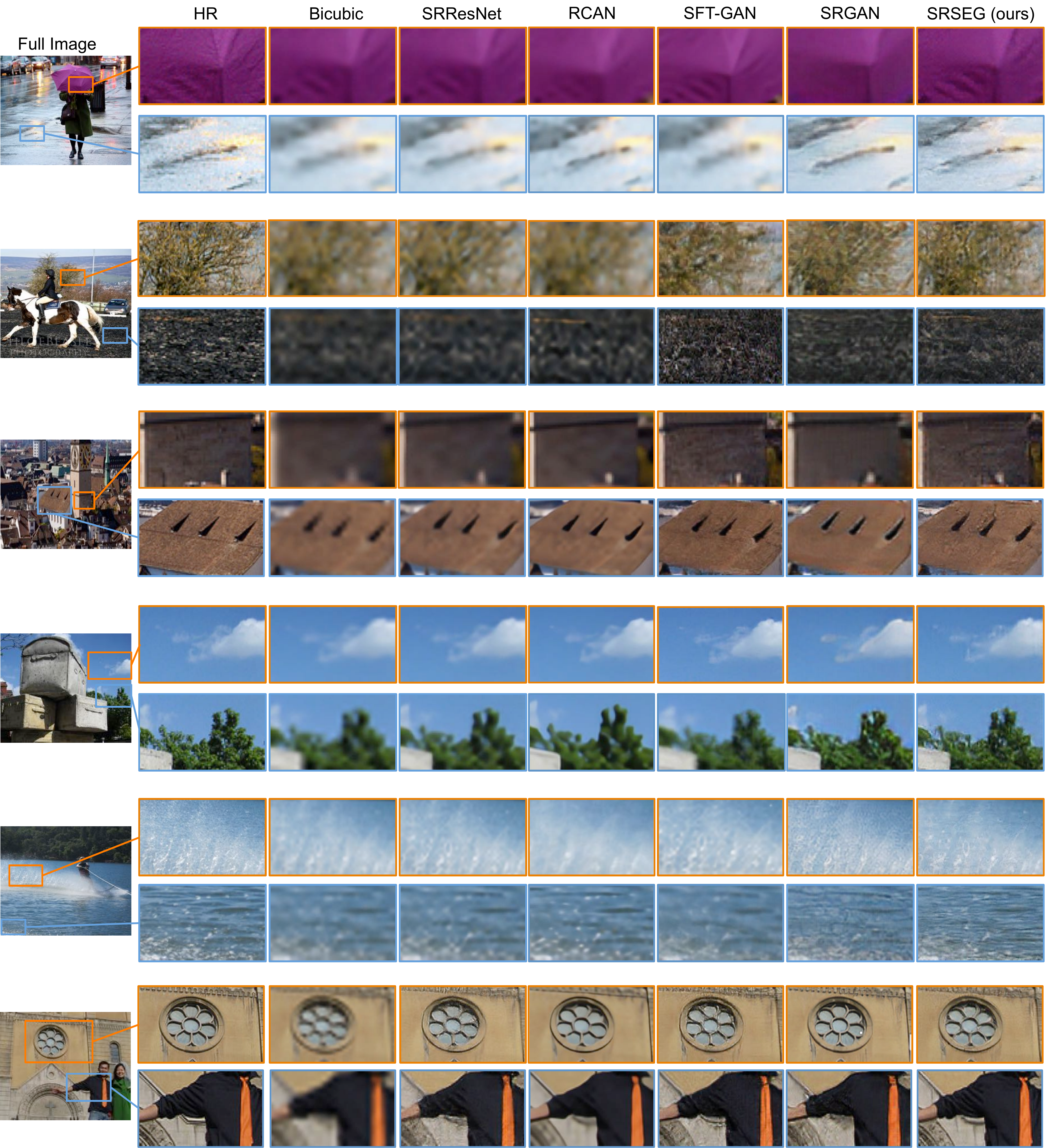}
\caption{Qualitative results on COCO-stuff dataset \cite{paper_coco_stuff}, focusing on object/background textures. The test images include images with the same categories as the one used during training (water, plant, building, sky, and ground). Cropped regions are zoomed in with a factor of 5 to 10. Images from left to right: High resolution image, bicubic interpolation, SRResNet \cite{paper_tweeter_0}, RCAN~\cite{paper_rcan}, SFT-GAN~\cite{paper_segmentation}, SRGAN \cite{paper_tweeter_0}, and SRSEG (this work). Zoom in to have the best view.}
\label{fig:segmentation_res_2}
\end{figure}

Figure \ref{fig:segmentation_res_2} contains visual examples comparing different models. In order to have a fair comparison, we re-trained the SRResNet~\cite{paper_tweeter_0}, SFT-GAN \cite{paper_segmentation}, and SRGAN \cite{paper_tweeter_0} methods on the same dataset and with the same parameters as ours. The generator and discriminator networks used in both SRGAN and our method are very similar (only layers resulting in segmentation probability output differ), which helps to investigate the effectiveness of our approach compared to the SRGAN, as the baseline. For RCAN, we used their pre-trained models in~\cite{paper_rcan}. The MATLAB imresize function with a bicubic kernel is used to produce LR images.

The qualitative comparison shows that our method generates more realistic and natural textures by benefiting from categorical information. Our experiment shows that the trained model for both segmentation and SR tasks is generalized in a way that it reconstructs more realistic background compared to the approaches using the same configuration and without the segmentation objective.

\begin{figure}
\centering
\includegraphics[height=2.5cm]{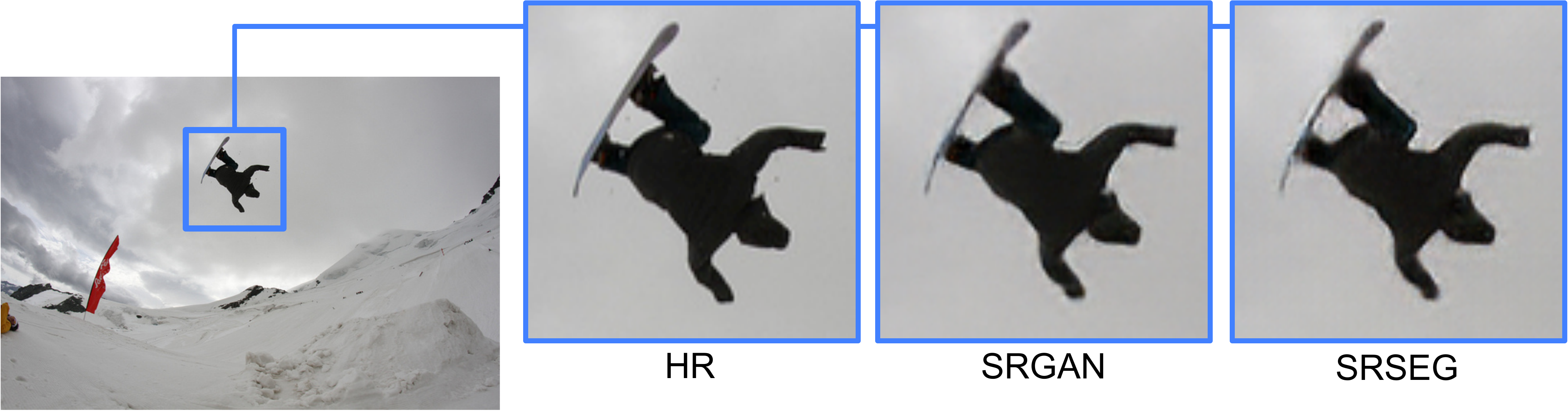}
\caption{An example of a bad reconstruction of boundaries compared to the SRGAN \cite{paper_tweeter_0} method; this effect could be seen in some cases, specially in objects/backgrounds that have not been from training classes.}
\label{fig:result_bad}
\end{figure}

As mentioned previously, to prove the concept, most of the test images contains specific background categories, however, it still reconstructs competitive results for objects without any labels during the training phase, e.g., the man with a tie in Figure~\ref{fig:segmentation_res_2}. In some cases, we could also observe that our method can result in a less precise boundaries as shown in Figure \ref{fig:result_bad}.

\subsection{User experience}

As \cite{paper_tweeter_0,paper_mse_0,paper_segmentation} mentioned, the commonly used quantitative measurements for SR methods, such as SSIM and PSNR, are not directly correlated to the perceptual quality; their experiments show that GAN-based methods have lower PSNR and SSIM values compared to PSNR-oriented approaches, however, they easily outperform them in terms of more appealing and closer images to the HR images. Therefore, we did not use these evaluation metrics in this work.

To better investigate the effectiveness of multitask learning simultaneously for semantic segmentation and SR, we perform a user study to compare the SRGAN \cite{paper_tweeter_0} method and our approach which is a an extended version of SRGAN with an additional segmentation output. We design our experiment in two stages; first stage quantifies the ability of our approaches to reconstruct perceptually convincing images while we focus specifically on the quality of texture reconstruction regarding to ground-truth (real HR image). 

\begin{figure}
\centering
\subfloat[]{{\includegraphics[height=2.8cm]{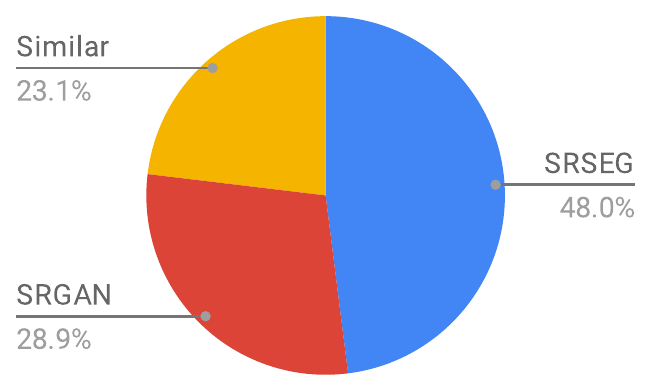} }}
\quad \quad \subfloat[]{{\includegraphics[height=2.8cm]{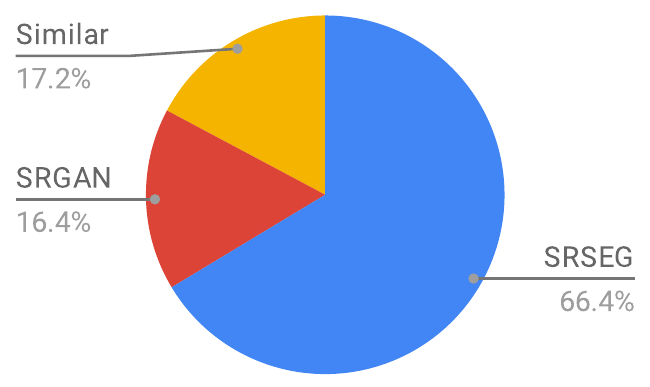} }}
\caption{The evaluation results of our user studies, comparing SRSEG (our method) with SRGAN \cite{paper_tweeter_0}; (a) Focusing on visual quality of the resolved images, (b) Focusing only on enlarged textures. Both textures and overall qualities of resolved images resolved by our method are improved. Users prefer textures reconstructed by our proposed approach by a large margin.}
\label{fig:pie_charts}
\end{figure}

During the first stage, users were requested to vote for more appealing images between SRGAN and our proposed method, SRSEG output pairs. In order to avoid random guesses in case of similar qualities, a third choice as "Similar" was also introduced for each image. 22 persons have participated in this experiment. 25 random images from COCO-Stuff \cite{paper_coco_stuff} were presented in a randomized fashion to each person. The pie chart shown in Figure~\ref{fig:pie_charts}.a illustrates that the images reconstructed by our approach are more appealing to the users.

In the second stage, we focused only on enlarged texture patches, zoomed in with a factor of 8 to 10, mostly on parts of backgrounds that have been from training classes. The enlarged images represent only a reconstructed texture and no object was included in the image. The ground-truth was also shown to users. Each person was asked again to pick the texture closer to the ground-truth. 25 pairs of textures in addition to their ground-truth were shown to 22 persons in this stage. The results of this stage is shown in Figure \ref{fig:pie_charts}.b. These results confirm that our approach reconstructs perceptually more convincing images for the users in terms of both overall and texture qualities of resolved images. However, comparing the results of the first and second stage of the user study shows that texture reconstruction quality of our proposed approach is by a large margin better than the quality of its object reconstruction. As a future work, adding more object categories to the training examples for both segmentation and SR tasks could also improve the reconstruction quality of the class ``Others'' with a similar margin.

\subsection{Ablation study}
\begin{figure}
\centering
\includegraphics[height=5.2cm]{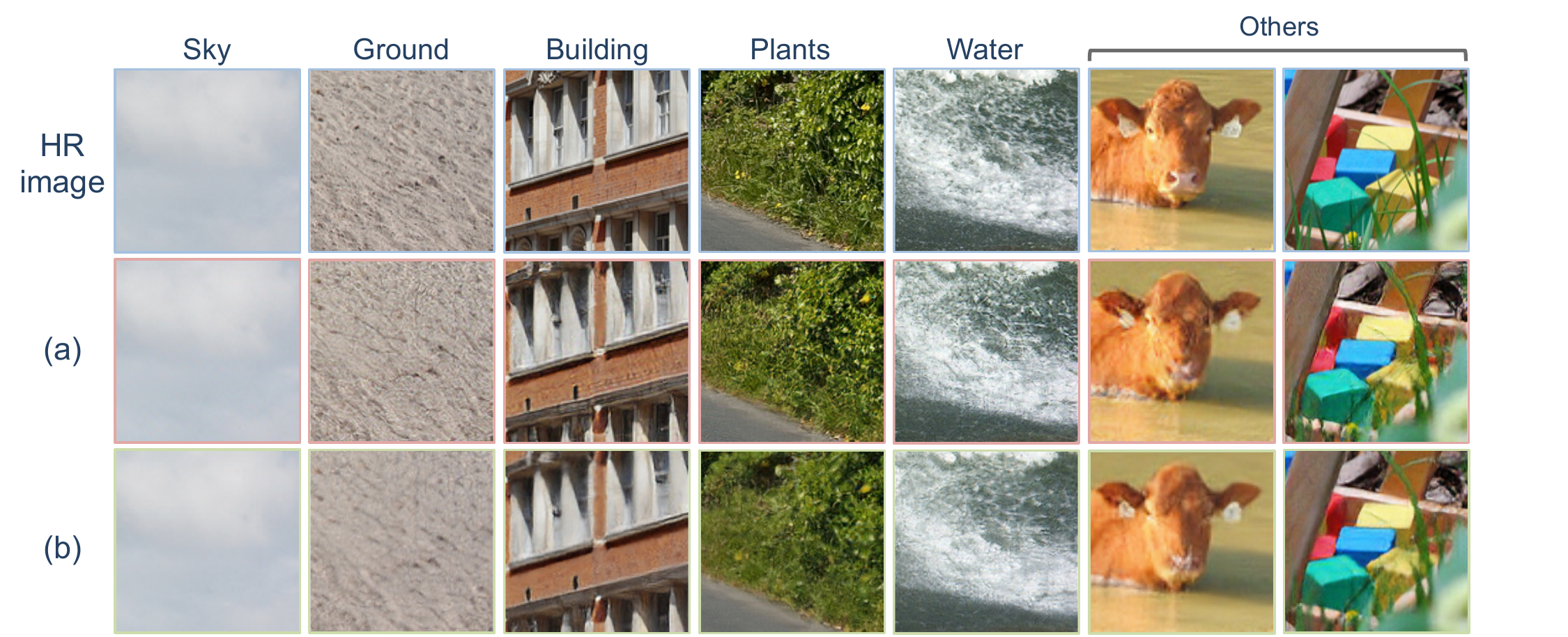}
\caption{Ablation study on different type of objects/backgrounds; comparing the reconstruction quality of our decoder: (a) with the segmentation extension during training, (b) without the segmentation extension. Zoom in for best view}
\label{fig:result_ablation}
\end{figure}

Intuitively, by introducing additional segmentation task, our SR decoder extracts more specific features for both image reconstruction and semantic segmentation. To investigate the competence of these new features and the effectiveness of our approach for image SR, we perform an ablation study, by qualitatively comparing the reconstruction quality of our decoder, with and without the segmentation extension. In Figure~\ref{fig:result_ablation}, we divide our results into different existing categories during training (sky, ground, buildings, plants, and water), as well as undefined categories in our dataset. We can see that the network trained with segmentation extension generates more photo-realistic textures for the available segmentation categories, while  having  competitive results for the other objects.

\subsection{Results on standard benchmarks}
\begin{figure}
\centering
\includegraphics[height=7.0cm]{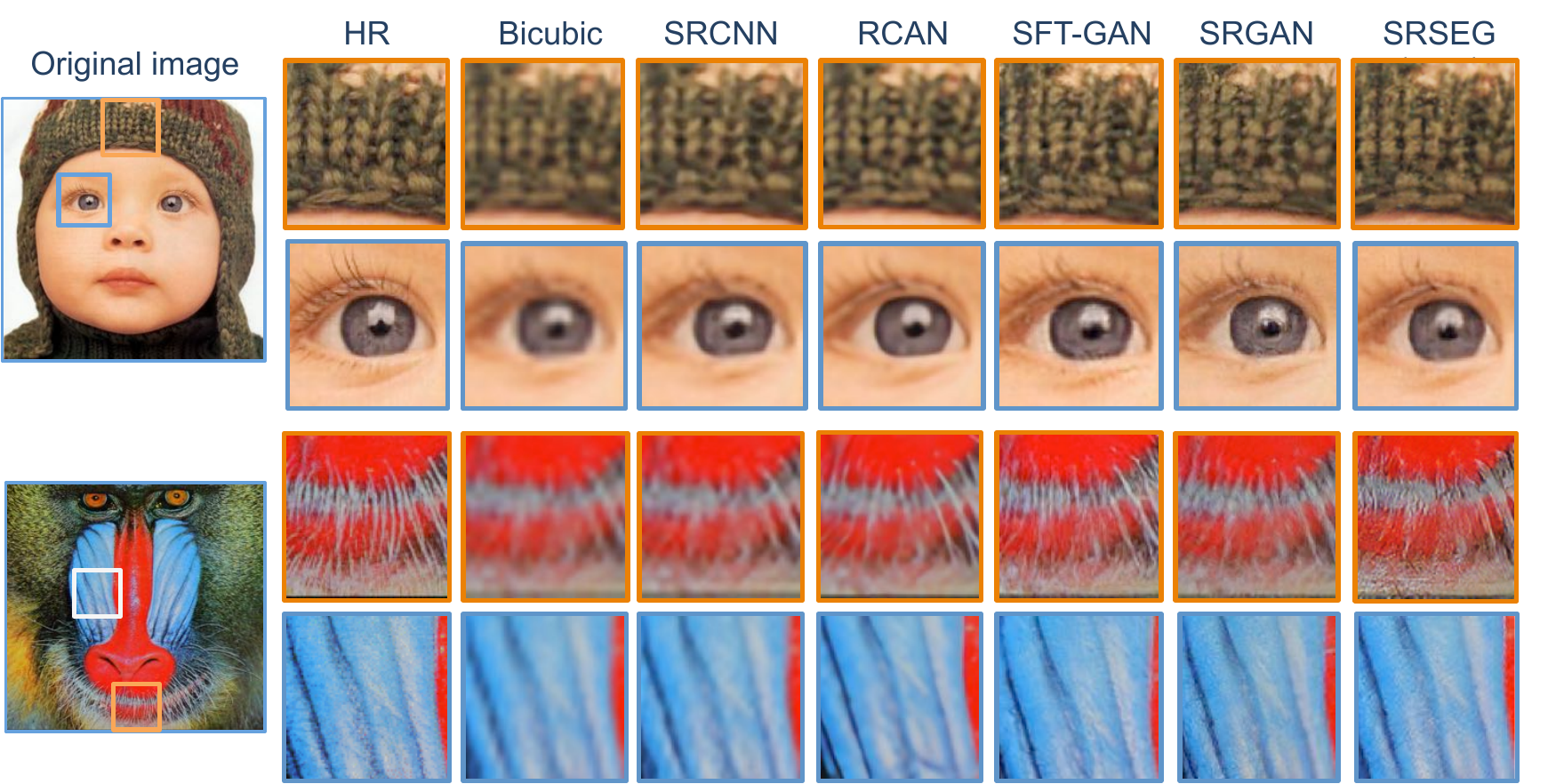}
\caption{Sample results on the “baby” (top) and “baboon” (bottom) images from Set5~\cite{paper_set5} and Set14~\cite{paper_set5} datasets, respectively. From left to right: HR image, bicubic, SRCNN~\cite{dong2016}, RCAN~\cite{paper_rcan}, SFT-GAN~\cite{paper_segmentation}, SRGAN~\cite{paper_tweeter_0}, and SRSEG (ours). Zoom in for the best view.}
\label{fig:result_set5set14}
\end{figure}

During training, our approach focuses on optimizing the decoder by using an additional segmentation extension and loss term for recognizing specific categories, such as sky, ground, buildings, plants, and water. Even though many object and background categories are absent during the training phase, our experiment shows that the model generalizes in a way that it reconstructs either more realistic or competitive results for undefined objects/backgrounds as well. In this section, we evaluate the reconstruction quality of unknown objects, by using Set5~\cite{paper_set5} and Set14~\cite{paper_set5} standard benchmarks, where unlike our training set, in most of the images, outdoor background scenes are not present. Figure~\ref{fig:result_set5set14} compares the results of our SR model on the ``baby'' and the ``baboon'' images to recent state-of-the-art methods including bicubic,  SRCNN~\cite{dong2016}, RCAN~\cite{paper_rcan}, SFT-GAN~\cite{paper_segmentation}, and SRGAN ~\cite{paper_tweeter_0}. In  both images, despite the fact that their categories were not existed during training, we could generate more photo-realistic images compared to SRCNN and RCAN, while  having competitive results with SFT-GAN and SRGAN. Their results were obtained by using their online supplementary materials.

\subsection{Computational time}
Our proposed method has similar running time to CNN-based SISR methods and faster than method such as \cite{paper_segmentation}, which uses a second network to predict segmentation probabilities. As the additional extension for segmentation, presented in this work, is removed at run-time and no segmentation label is required as an input, the running time is not affected by our proposed approach. However, using segmentation extension during the training phase increases our training time with a factor of $1.3$ compared to SRGAN. 

In particular, our Tensorflow implementation runs at 20.24 FPS on a GeForce GTX 1080 Ti graphic card to reconstruct HD images ($1024 \times 768$) from their low-resolution counter-parts ($256 \times 192$) with a scale factor of 4.

\section{Conclusion and Future Work}
\label{sec:conclusion}
In this work we presented a novel approach to use categorical information to tackle the SR problem. We introduced a SR decoder only benefiting from one shared deep network to learn simultaneously image super-resolution and semantic segmentation by keeping two task-specific output layers during training. We also introduced a novel boundary mask to filter out unrelated segmentation losses caused by imprecise segmentation labels. We have conducted perceptual experiments including a user study on images from COCO-Stuff dataset and demonstrated that multitask learning can enable benefiting from semantic information in a single network and improves the recovering quality. As a future work, additional object/background categories can be introduced during the training in order to explore how it could affect the reconstruction quality.\\

\setlength{\parindent}{0cm}
\textbf{Acknowledgement} 
This work was funded and supported by Swisscom Digital Lab in Lausanne, Switzerland.

\bibliographystyle{model1-num-names}
\bibliography{my_bibliography.bib}

\end{document}